\documentclass[10pt,twocolumn,letterpaper]{article}

\usepackage{cvpr}
\usepackage{times}
\usepackage{epsfig}
\usepackage{graphicx}
\usepackage{amsmath}
\usepackage{amssymb}
\usepackage{bbm}
\usepackage{lib}
\usepackage{caption}
\usepackage{mathtools}
\usepackage{multirow}
\usepackage{epigraph}
\usepackage{booktabs}
\usepackage{balance}

\usepackage[urlcolor=red,pagebackref=true,breaklinks=true,letterpaper=true,colorlinks,bookmarks=false]{hyperref}

\cvprfinalcopy 

\def\cvprPaperID{000} 

\begin{document}

\title{Learning Shape Abstractions by Assembling Volumetric Primitives}

\author{Shubham Tulsiani$^1$, Hao Su$^2$, Leonidas J. Guibas$^2$, Alexei A. Efros$^1$, Jitendra Malik$^1$  \\
$^1$University of California, Berkeley ~~~~ $^2$Stanford University\\
{\tt\small$^1$\{shubhtuls, efros, malik\}@eecs.berkeley.edu,  $^2$\{haosu, guibas\}@cs.stanford.edu}
}

\twocolumn[{%
\renewcommand\twocolumn[1][]{#1}%
\vspace{-1em}
\maketitle
\vspace{-1em}
\begin{center}
   \centering \includegraphics[width=\textwidth]{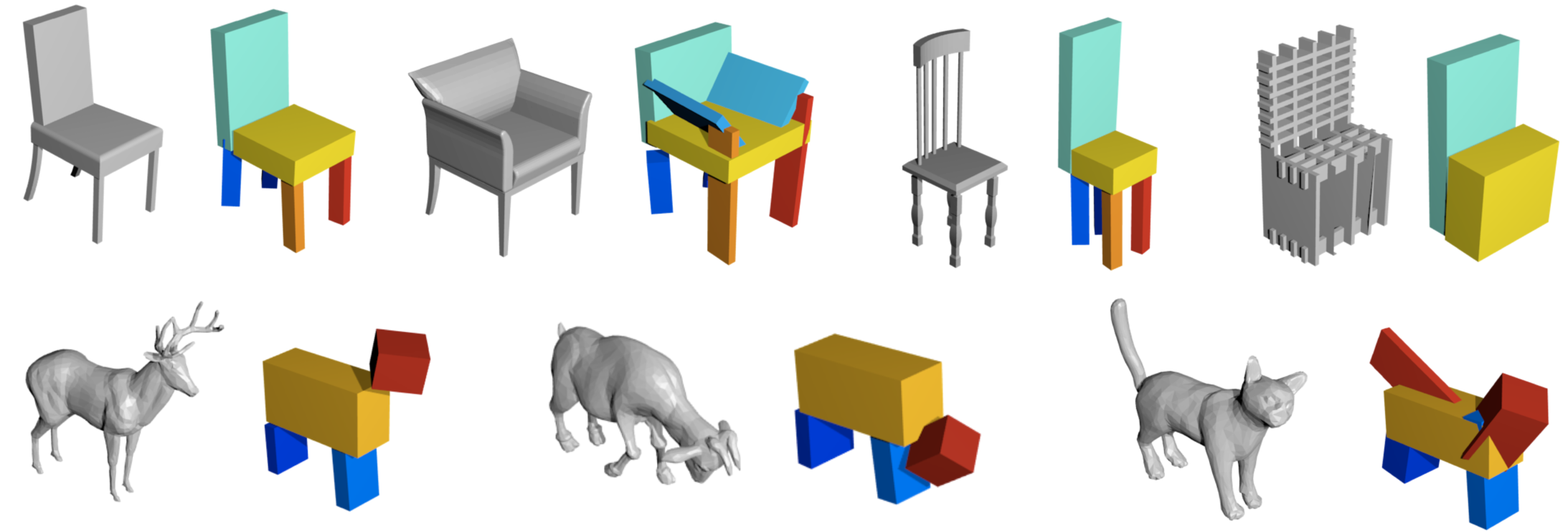} \captionof{figure}{Examples of chair and animal shapes assembled by composing simple volumetric primitives (cuboids). The obtained reconstructions allows an interpretable representation for each object and provides a consistent parsing across shapes \eg chair seats are captured by the same primitive across the category.}
   \figlabel{fig1}
\end{center}%
}]


\begin{abstract}
We present a learning framework for abstracting complex shapes by learning to assemble objects using 3D volumetric primitives. In addition to generating simple and geometrically interpretable explanations of 3D objects, our framework also allows us to automatically discover and exploit consistent structure in the data.  We demonstrate that using our method allows predicting shape representations which can be leveraged for   obtaining a consistent parsing across the instances of a shape collection and constructing an interpretable shape similarity measure. We also examine applications for image-based prediction as well as shape manipulation.
\end{abstract}


\section{Introduction}

\vspace{-2mm}
\epigraph{``Treat nature by means of the cylinder, the sphere, the cone, everything brought into proper perspective"}{\textit{Paul Cezanne}}

Cezanne's insight that an object can be conceived as assembled from a set of volumetric primitives has resurfaced multiple times in the vision and graphics literature. In computer vision, generalized cylinders were introduced by Binford back in 1971, where a cross-sectional area is swept along a straight or curved axis while possibly being shrunk or expanded during the process~\cite{binford1971}. One of the key motivations was {\em parsimony of description} -- an object could be described by relatively few generalized cylinders, each of which in turn requiring only a few parameters. Volumetric primitives remained popular through the 1990s  as they provided a coherent framework for explaining shape inference from a single image, perceptual organization, as well as  recognition of a 3D object from 2D views.  However, fitting generalized cylinders to image data required considerable hand crafting, and as machine learning techniques for object recognition came to the fore in the 1990s, 
this paradigm faded from the main stage.


Of course, finding parsimonious explanations for complex phenomena lies at the core of learning-based visual understanding. Indeed, machine learning is only possible because our visual world, despite its enormous complexity, is also highly structured -- visual patterns don't just happen once, but keep on repeating in various configurations. In contemporary computer vision, this structure is most often modeled via human supervision: the repeating patterns are labeled as objects or object parts, and supervised learning methods are employed to find and name them in novel imagery.  However, 
it would seem more satisfying if complex structures could be explained in terms of simpler underlying structures. 


In this paper we return to the classic problem of explaining objects with volumetric primitives, but using the modern tools of unsupervised learning and convolutional neural networks (CNNs).  We choose the simplest possible primitives, rigidly transformed cuboids, and show how deep convolutional networks can be trained to assemble arbitrary 3D objects out of them (at some level of approximation). The main reason we succeed where the classic approaches failed is because we aim to explain the entire dataset of 3D objects jointly, allowing us to learn the common 3D patterns directly from the data.  

While the representation of the 3D object shapes \eg as meshes or voxel occupancies, is typically complex and high-dimensional, the resulting explanation in terms of basic primitives is parsimonious, with a small number of parameters. 
As examples of their  applicability, we leverage the primitive based representation for various tasks \eg part discovery, image based abstraction, shape manipulation \etc. Here we do not wish to reprise the classic debates on the value of volumetric primitives -- while they were oversold in the 70s and 80s, they suffer from complete neglect now, and we hope that this demonstration of feasibility of learning how to assemble an object from volumetric primitives will reignite interest. Code is available at \href{https://shubhtuls.github.io/volumetricPrimitives}{https://shubhtuls.github.io/volumetricPrimitives}.

\section{Related Work}

\vspace{2mm}
\noindent \textbf{3D Representation and Reconstruction.} 
The classic approaches for modeling objects and scenes dating to the very beginnings of the computer vision discipline, such as blocks world~\cite{roberts1963machine},
generalized cylinders~\cite{binford1971}, and geons~\cite{biederman1987recognition}, emphasized the compactness of representation as the central goal.  In a similar spirit, a few modern approaches have attempted to reconstruct objects/scenes using simple primitives, including Lego pieces~\cite{van2015part} and qualitative 3D blocks~\cite{gupta2010blocks}.
Apart from these attempts, most mainstream methods for representing and reconstructing objects typically use much higher-dimensional representations \eg objects as point clouds~\cite{shapesKarTCM15, vicente2014reconstructing} or exemplar CAD models~\cite{li2015joint, limparsing, pascal3d}. The success of the latter set of approaches has been largely driven by the data-driven reasoning which the classical methods did not leverage. Our work aims to combine the two -- we aim for a parsimonious representation but discover the underlying parsimony in a data-driven manner instead of relying on hand-crafted cues and priors. Similar to our approach, Yumer and Kara~\cite{yumer2012co, yumer2014co} showed that parsimonious modelling with data-driven reasoning can allow consistent geometry simplifications or deformations in shape collections but our learning based approach allows efficient test time inference for novel shapes. An additional  property of our approach, compared to classical methods, is the consistency of representation across instances. Classical approaches solve a per-instance optimization and obtain an \emph{unordered set of primitives} whereas our our approach outputs a consistent \emph{indexed set of primitives} -- this allows several applications examined in \secref{applications}.

\vspace{2mm}
\noindent \textbf{Parsing Objects, Scenes and 3D Shapes.} 
The idea of exploiting repeating structures in large datasets has been central to efforts on unsupervised object discovery and co-segmentation~\cite{rubinstein2013, russell2006using}.  Data-driven compositionality, in particular, has been used for co-segmentation~\cite{faktor2013}, scene parsing and novel scene generation~\cite{SceneCollaging, russell2009segmenting}.
In the domain of 3D shapes, the idea of exploiting compositionality has played a similarly important role for object representation, parsing, and manipulation. Pre-labeled, part-based shape representations were used for capturing the category-specific shape manifold~\cite{fakscm_metarep_sig14}, generating novel objects~\cite{huang2015single, kalogerakis2012} or recovering 3D from 2.5D data~\cite{sung2015data}. Other methods aim to automatically discover these components in 3D shape datasets~\cite{huang2011joint}, and their relative arrangements~\cite{zcam_partArrangement_eg14}. Similar to these shape and  scene based methods, our framework can automatically discover consistent components and understand the structure of the data, but we do so by virtue of learning to generate parsimonious explanations.

\vspace{2mm}
\noindent \textbf{Deep Generative Models.}
The rapid recent progress in supervised learning tasks by using deep learning techniques has been accompanied by a growing interest in leveraging similar methods to discover structure in the visual data. Using generative adversarial networks~\cite{goodfellow2014generative,radford2015unsupervised} allows learning the data distribution but the underlying latent space lacks interpretability. Other generative methods aim to explicitly decouple the underlying factors of variation~\cite{cheung2014discovering, kulkarni2015deep} but rely on supervision for disentangling these factors. More closely related to our work, some recent approaches use recurrent networks to iteratively generate components to explain a simple 2D input scene~\cite{eslami2016attend, gregor2015draw, huang2015efficient}. Our work uses similar principles of learning component based explanations of complex shapes where the components are interpretable simple 3D primitives.

\begin{figure*}[t!]
\centering
\includegraphics[width=1.00\textwidth]{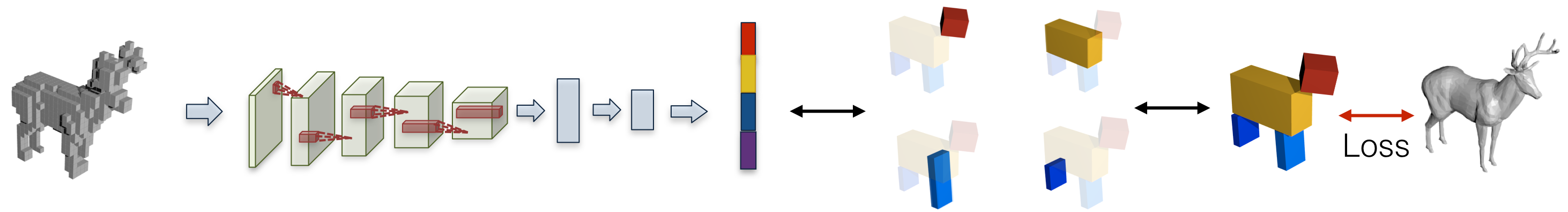}
\caption{Overview of our approach. Given the input volume corresponding to an object $O$, we use a CNN to predict primitive shape and transformation parameters $\{(z_m,q_m,t_m)\}$ for each part (\secref{primRep}). The predicted parameters implicitly define transformed volumetric primitives $\{\bar{P}_m\}$ whose composition induces an assembled shape. We train our system using a loss function which attempts to minimize the discrepancy  between the ground-truth mesh for $O$ and the assembled shape which is implicitly defined by the predicted parameters (\secref{primLoss}).}
\figlabel{framework}
\end{figure*}

\section{Learning Object Assembly}
\seclabel{assembly}

We formulate the problem of assembling a target object $O$, given input signal $I$ as that of predicting (up to) $M$ distinct parts which are then composed to output the final shape. Towards this, we learn a CNN $h_{\theta}$ parametrized by $\theta$ which outputs a primitive based representation. The task of learning this CNN is an unsupervised one -- we do not  have any annotations for the primitive parameters that best describe the target objects. However, even though there is no direct supervision, one can measure if a predicted primitive configuration is good by checking if the assembled object matches the target object. Using this insight, we formulate a loss function which informs us if the shape assembled using the predicted primitives matches the target shape and optimize this loss to train the CNN.

An overview of our approach is presented in \figref{framework}. Given a discretized representation of the target shape as input, we use a CNN to predict a primitive representation (described in \secref{primRep}). The predicted representation implicitly defines an assembled shape by composing the predicted primitives. \secref{primLoss} describes a differentiable loss function that allows using this representation in a learning framework.  While the initial presentation assumes the use of a fixed number of primitives, \secref{varNum} extends our approach to allow a variable number of primitives.



\subsection{Primitive based Representation}
\seclabel{primRep}
We represent an assembled shape by composing the predicted simple transformed primitives. Each primitive is encoded in terms of a tuple $(z,q,t)$ where $z$ represents its shape in a canonical frame and  $(q,t)$ represent the spatial transformation (rotation and translation). The assembled shape predicted by the neural network  $h_{\theta}$ can therefore be written as below.

\begin{gather}
\eqlabel{netPred} \{(z_m, q_m, t_m) | m~=~1, \cdots, M\} = h_{\theta}(I)
\end{gather}

The motivation for this parametrization is to exploit the compositionality of parts as well as the independence of `what' and `where' (part shape and spatial transformation respectively). The representation of a shape as a set of parts allows independent reasoning regarding semantically separate units like chair legs, seat \etc. The decomposition in terms of part shape and transformation parameters further decomposes factors of variation like `broad aeroplane wing' (captured by shape) and `tilted chair back' (captured by transformation).


\subsection{Loss Function for Assembled Shape}
\seclabel{primLoss}
We want to define a differentiable loss function $L(\{(z_m, q_m, t_m)\},O)$ between the CNN prediction $\{(z_m, q_m, t_m)\}$ and the target object $O$. This is a challenging task because the prediction and the groundtruth have different 3D representations -- the prediction is a parametrized shape whereas the groundtruth is a mesh consisting of triangles. To overcome this, we leverage the fact that the parametrization in terms of simple primitives allows efficient computation of some properties of the shape induced by their composition. In particular, we can compute the \textit{distance field} (\secref{prelim}) of the assembled shape as well as sample points on the surface of the primitives. These allow us to define two complimentary losses  which together aim to minimize the discrepancy between the predicted and ground-truth shape. The \textit{Coverage Loss} tries to enforce that the object $O$ is subsumed by the predicted assembled shape. The \textit{Consistency Loss} enforces the other direction -- that the object $O$ subsumes the predicted shape. By optimizing these losses together, we ensure that the assembled shape tries to be maximally consistent with the target object.


%



\subsubsection{Preliminaries}
\seclabel{prelim}
\vspace{2mm}
\noindent \textbf{Notation.}
We represent by $P_m$, the \textit{untransformed primitive} as predicted according to $z_m$ and use $\bar{P}_m$ to denote the primitive $P_m$  after rotation, translation according to $(q_m, t_m)$. Therefore, the final shape induced by the composition of the predicted primitives is $\underset{m}{\cup} \bar{P}_m$.

We use the function $S(\cdot)$ to represent the surface of the argument and $p \sim S(\cdot)$ represents  a random point sampled on it \eg $p \sim S(\bar{P}_m)$ corresponds to a point sampled on the surface of $m^{th}$ primitive. We also require notations for simple rigid transformations -- we denote by $\mathcal{R}(p, q)$ result of rotating a point $p$ according to rotation specified by quaternion $q$ and similarly, $\mathcal{T}(p, t)$ denotes the result of translating a point $p$ by $t$. Note that the operations $\mathcal{R}, \mathcal{T}$ are both differentiable.

\vspace{2mm}
\noindent \textbf{Distance Field.} A distance field $\mathcal{C}(~\cdot~; O)$ corresponding to an object $O$
is a function $\mathbb{R}^3 \rightarrow \mathbb{R^+}$ that computes the distance to the closest point of the object. Note that it evaluates to 0 in the object interior.
\begin{gather}
\mathcal{C}(p; O) =  \underset{p' \in O}{min}~\|p - p' \|_2
\end{gather}

\subsubsection{Coverage Loss : $O \subseteq \underset{m}{\cup} \bar{P}_m$ .}
\seclabel{coverage}
We want to penalize the CNN prediction if the target object $O$ is not completely covered by the predicted shape  $\underset{m}{\cup} \bar{P}_m$. A sufficient condition to ensure this is that the distance field of the assembled shape evaluates to zero for all points on the surface of O.
\begin{gather}
L_1(\{(z_m, q_m, t_m)\},O) = \mathbb{E}_{p \sim S(O)} \|  \mathcal{C}(p; \underset{m}{\cup} \bar{P}_m) \|^2
\eqlabel{coverageLoss}
\end{gather}

Computation can be simplified due to a nice property of distance fields. It is easy to show that the distance field of a composed shape equals to the pointwise minimum of the distance fields of all composing shapes:
\begin{gather}
\mathcal{C}(p; \underset{m}{\cup} \bar{P}_m) = \underset{m}{min} ~ \mathcal{C}(p; \bar{P}_m)
\eqlabel{distDecomp}
\end{gather}

This decomposition rule boils the distance field of a whole shape down to the distance field of a primitive. In the following, we show how to efficiently compute $\mathcal{C}$ for primitives as cuboids.

\vspace{2mm}
\noindent \textbf{Distance field of Primitives.}
Given an origin-centred cuboid represented by $z \equiv (w, h, d)$ -- its extent in the three dimensions, its distance field $\mathcal{C}_{cub}(~\cdot~ ; z)$ can be computed as below (using $\text{max}(0,x) \equiv x_+$):
\begin{gather*}
\mathcal{C}_{cub}(p ; z)^2 = (|p_x| - w)_+^2 +  (|p_y| - h)_+^2 +  (|p_z| - d)_+^2
\end{gather*}

Consider an object $O$ (with an associated field $\mathcal{C}(~\cdot~; O)$) undergoing a rotation $R$ (parametrized by quaternion $q$) followed by a translation $t$. The distance field at a point $p$ w.r.t. the transformed object is the same as the distance field at $p'$ wrt. the canonical object where $p' = R^{-1}(p - t)$. This observations allows us to complete the formulation by defining $\mathcal{C}(p; \bar{P}_m)$ (required in \eqref{distDecomp}) as below.

\begin{gather}
\mathcal{C}(p; \bar{P}_m) = \mathcal{C}(p'; P_m) ; p' = \mathcal{R}(\mathcal{T}(p, -t_m),\bar{q}_m) \\
\mathcal{C}(~\cdot~; P_m) = \mathcal{C}_{cub}(~\cdot~; z_m)
\eqlabel{CPmcub}
\end{gather}

\subsubsection{Consistency Loss : $\underset{m}{\cup} \bar{P}_m \subseteq  O$.}
\seclabel{consistency}
We want to penalize the CNN prediction if the  predicted shape  $\underset{m}{\cup} \bar{P}_m$ is not completely inside the target object $O$. A sufficient condition is to ensure this is that the distance field of the object O shape evaluates to zero for all points on the surface of individual primitives $\bar{P}_m$.
\begin{gather}
L_2(\{(z_m, q_m, t_m)\},O) = \underset{m}{\sum} ~ \mathbb{E}_{p \sim S(\bar{P}_m)} \|  \mathcal{C}(p; O) \|^2
\eqlabel{ConsistencyLoss}
\end{gather}

Additionally, we observe that to sample a point $p$ on the surface of $\bar{P}_m$, one can equivalently sample $p'$ on the surface of the untransformed primitive $P_m$ and then rotate, translate $p'$ according to $(q_m, z_m)$.
\begin{gather*}
p \sim S(\bar{P}_m) \equiv \mathcal{T}(\mathcal{R}(p', q_m), t_m); p' \sim S(P_m)
\end{gather*}
An aspect for computing gradients for the predicted parameters using this loss is the ability to compute derivatives for $z_m$ given gradients for a sampled point on the canonical untransformed primitive $p' \sim S(P_m)$. We do so by using the \textit{re-parametrization trick}~\cite{kingma2013auto} which decouples the parameters from the random sampling. As an example, consider a point being sampled on a rectangle extending from $(-w,-h)$ to $(w,h)$. Instead of sampling the x-coordinate as $x \sim [-w,w]$, one can use $u \sim [-1,1]$ and $x = uw$. This re-parametrization of sampling allows one to compute $\frac{\partial x}{\partial w}$. We provide the details for applying the re-parametrization trick for a cuboid primitive in the appendix.


\subsection{Allowing Variable Number of Primitives}
\seclabel{varNum}
The framework we have presented so far reconstructs each instance in an object category using exactly $M$ primitives. However, different instances in an object category can be explained by different number of primitives \eg some chairs have handles, others don't. To incorporate this, in addition to predicting the shape and transformation of each primitive, we also predict the probability of its existence $p_m$. We first discuss the modified representation predicted by the CNN and discuss how the loss function can  incorporate this.

\vspace{2mm}
\noindent \textbf{Primitive Representation.} 
As we mentioned above, the primitive representation has an added parameter $p_m$ -- the probability of its existence. To incorporate this, we factor the primitive shape $z_m$ into two components -- $(z_m^s, z_m^e)$. Here $z_m^s$ represents the primitive's dimensions (\eg cuboid height, width, depth) as before and $z_m^e \sim Bern(p_m)$ is a binary variable which denotes if the primitive actually exists \ie if $z_m^e = 0$ we pretend as if the $m^{th}$ primitive does not exist. The prediction of the CNN in this scenario is as below.
\begin{gather}
\eqlabel{netPredStoch} \{(z_m^s, q_m, t_m, p_m) | m~=~1 \cdots M\} = h_{\theta}(I) \\
\forall _m ~ z_m^e \sim Bern(p_m) ; ~ z_m \equiv (z_m^s, z_m^e)
\end{gather}

Note that the CNN predicts $p_m$ -- the parameter of the Bernoulli distribution from which the part existence variable $z_m^e$ is sampled. This representation allows the prediction of a variable number of parts \eg if a chair is best explained using $k < M$ primitives, the network can  predict a high $p_m$ for only $k$ primitives and a low  $p_m$ for the remaining  $M-k$ primitives.

\vspace{2mm}
\noindent \textbf{Learning.}
Under the reformulated representation of primitives, the CNN output does not induce a unique assembled shape -- it induces a distribution of possible shapes where the $m^{th}$ primitive stochastically exists with probability $p_m$. In this scenario, we want to minimize the expected loss across the possible assemblies. The first step is to modify the consistency and coverage losses to incorporate $z_m \equiv (z_m^s, z_m^e)$. Towards this, we note that the untransformed primitive $P_m$ is either a cuboid (if $z_m^e = 1$) or empty (if $z_m^e = 0$). In case it is empty, we can simply skip it the the consistency loss (\secref{consistency}) for this primitive and can incorporate this in the coverage loss (\secref{coverage}) by modifying \eqref{CPmcub} as follows -
\begin{gather}
\eqlabel{CPm_null} \mathcal{C}(~\cdot~; P_m) = 
\begin{dcases}
   \infty , & \text{if } z_m^e = 0\\
   \mathcal{C}_{cub}(~\cdot~; z_m^s),   & \text{if } z_m^e = 1\\
\end{dcases}
\end{gather}

We can now define the final loss $L(h_{\theta}(I),O)$ using the concepts developed. Note that this is simply the expected loss across possible samplings of $z_m^e$ according to $p_m$.

\begin{multline}
L(\{(z_m, q_m, t_m)\},O) = L_1(\{(z_m, q_m, t_m)\},O) \\ + L_2(\{(z_m, q_m, t_m)\},O)
\end{multline}
\begin{gather*}
L(h_{\theta}(I),O) = \mathbb{E}_{\forall m ~ z_m^e \sim Bern(p_m)}  L(\{(z_m, q_m, t_m)\},O)
\end{gather*}

Under this loss function, the gradients for the continuous variables \ie $\{ (z_m^s, q_m, t_m) \}$ can be estimated by averaging their gradients across samples. However, to compute gradients for the distribution parameter $p_m$, we use the REINFORCE algorithm \cite{williams1992simple} which basically gives positive feedback if the overall error is low (reward is high) and negative feedback otherwise. To further encourage parsimony, we include a small \textit{parsimony reward} (reward for choosing fewer primitives) when computing gradients for $p_m$.

\section{Experiments}
\begin{figure*}[t!]
\centering
\includegraphics[width=1.0\textwidth]{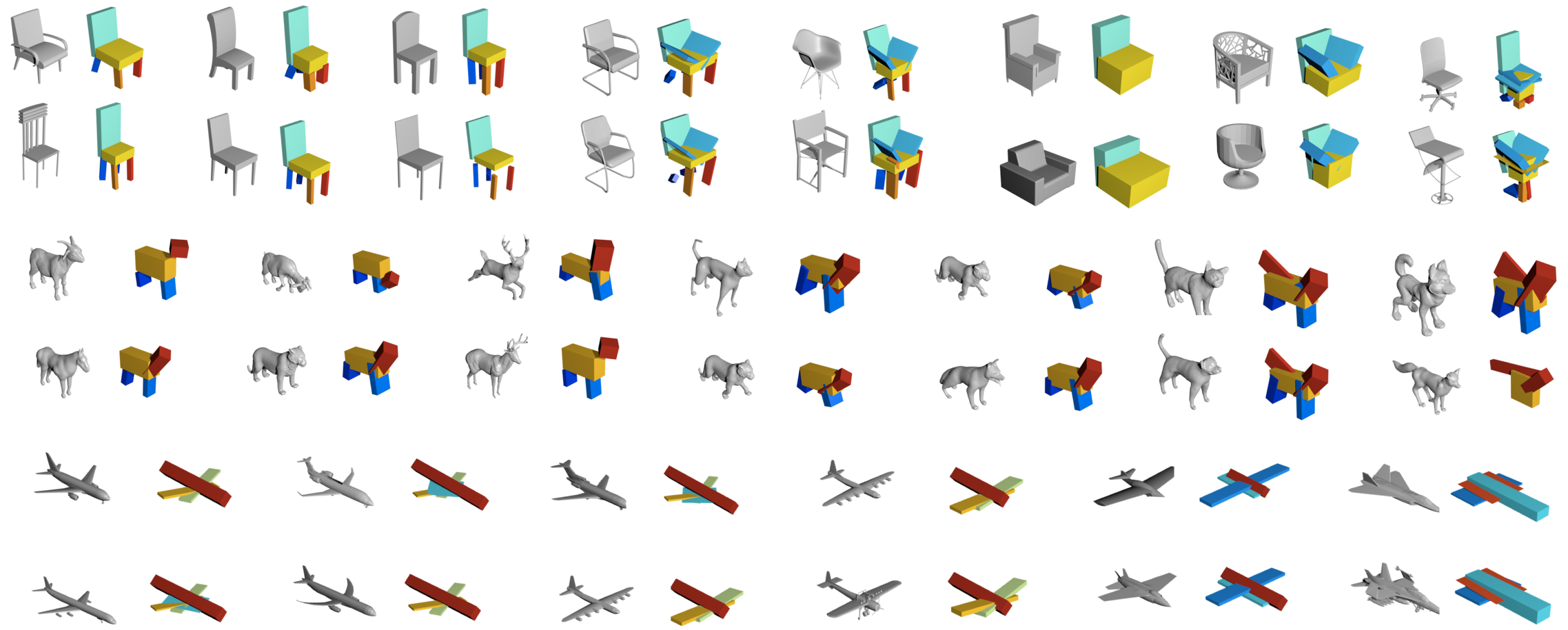}
\caption{Final predictions of our method on chairs, animals and aeroplanes. We visualize the more commonly occurring modes on the left and progressively towards the right show rarer configurations predicted.}
\figlabel{predFig}
\end{figure*}

\vspace{2mm}
\noindent \textbf{Dataset.} We perform our experiments primarily using the ShapeNet~\cite{shapenet2015} dataset which has a large collection of 3D models. In particular, we use the `airplane' and `chair' object categories which have thousands of meshes available. The ShapeNet models are already aligned in a canonical frame and are of a fixed scale. Additionally, in order to demonstrate applicability beyond rigid objects, we also manually download and similarly preprocess a set of around 100 models corresponding to four-legged animals.

\vspace{2mm}
\noindent \textbf{Network Architecture and Training.}
The dataset described above gives us a set of 3D objects $\{O_i\}$. Corresponding to $O_i$, the input to our CNN is a discretized  representation as a volumetric occupancy grid $I_i$ of size $32 \ast 32 \ast 32$ (we later experiment with rendered images as input in \secref{imgpred}). The encoder used in our shape assembler, as shown in \figref{framework}, takes in as input an occupancy grid and passes it through 3D convolutional and fully connected layers with intermediate non-linearities to output the primitive parameters  $\{(z_m^s, q_m, t_m, p_m) | m~=~1 \cdots M\} \equiv h_{\theta}(I_i)$. In this work, we use cuboid primitives and $z_m^s$ represents the width, height and thickness of cuboids. We use ADAM~\cite{adam} to train our network according to the loss $L(h_{\theta}(I_i), O_i)$ described in \secref{assembly} which aims to make the assembled shape predicted using $I_i$ match to the target object $O_i$. 

\begin{figure}[htb!]
\centering
\includegraphics[width=0.5\textwidth]{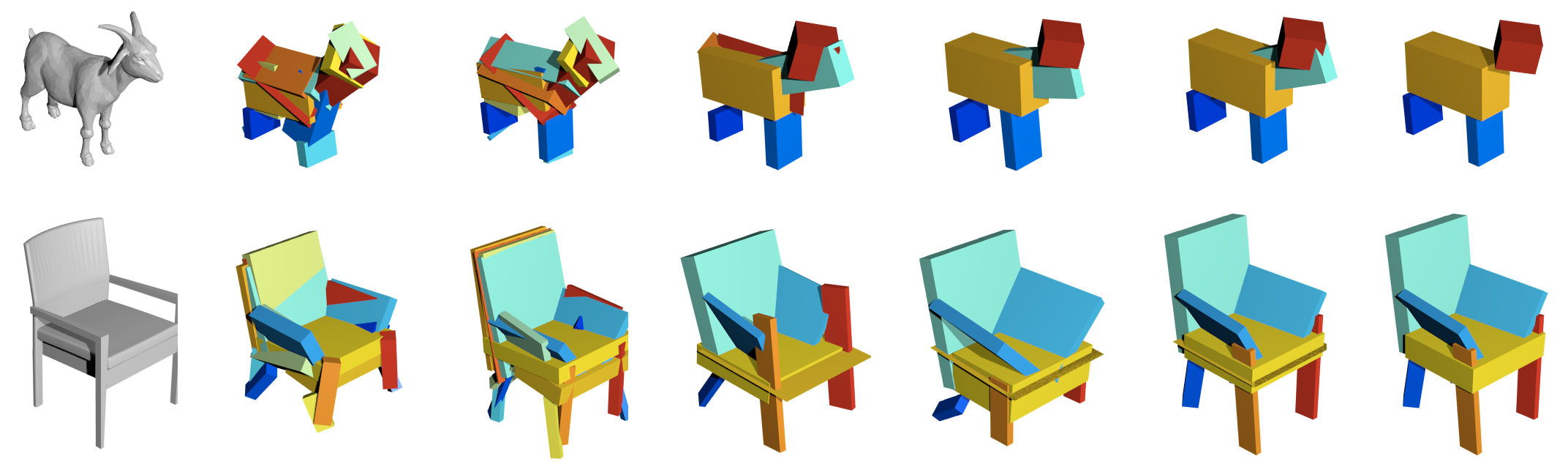}
\caption{Visualization of the training progression. We visualize the prediction for two instances (shown in column 1) after every 10,000 iterations (left to right, in columns 2-6). The last column shows the result after post-processing to remove redundant parts that overlap significantly with others. The initial training stage (up to 20,000 iterations) uses all primitives but we later allow the network to learn to use fewer primitives and the predictions gradually become more parsimonious.}
\figlabel{trainingIter}
\end{figure}

\begin{figure}[htb!]
\centering
\includegraphics[width=0.5\textwidth]{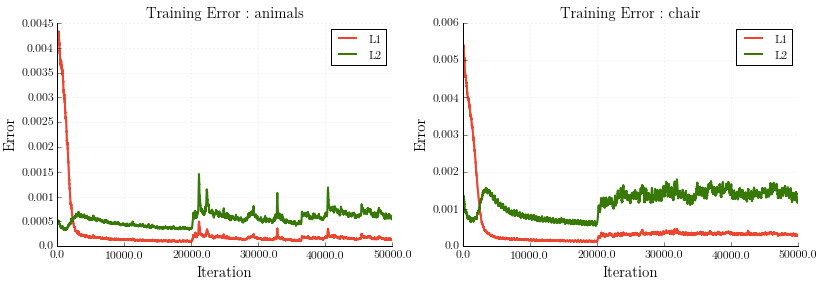}
\caption{We plot the Coverage (L1) and Consistency (L2) losses over training iterations. The losses both decreases in the initial stage of training (up to 20,000 iterations) but when we allow the use of varying number of primitives along with parsimony reward, the losses initially increase. This reveals a tradeoff between representation parsimony  and reconstruction accuracy.}
\figlabel{trainingErr}
\end{figure}

\vspace{2mm}
\noindent \textbf{Implementation Details.}
The coverage and consistency loss functions are both defined using expectations over sampled points. In practice, we randomly sample 1000 points on $S(O)$ to implement \eqref{coverageLoss} and 150 points from each $S(\bar{P}_m)$ to implement \eqref{ConsistencyLoss}. To efficiently compute the distance field of the target object $O$ at an arbitrary point $p$ in  \eqref{ConsistencyLoss}, we precompute the distance field and its derivatives for samples in a dense regular grid and use it to obtain efficient but  approximate gradients $\frac{\partial \mathcal{C}(p, O)}{\partial p}$.

Another practical difficulty  is that the gradients for the primitive existence probabilities $p_m$ are extremely noisy in the initial training stages --  \eg in the initial stages if a primitive is incorrectly placed, the CNN may learn to predict a very small $p_m$ instead of learning to align the primitive correctly. To overcome this, we use a two-stage training process. We first train the network using a fixed high value of $p_m$ across primitives and later allow the network to also learn $p_m$ while also encouraging simplicity by the external parsimony reward. As shown in  \figref{trainingErr}, this has the effect of first using a large number of primitives and in later stages, merging them together and using fewer primitives.

After the CNN has been trained, when computing the assembled representation for an object, we use MLE estimates instead of sampling \ie $z_m^e = \mathbbm{1}(p_m > 0.5)$. The final shape predictions using the CNN may still have redundant parts used and we use a simple post-processing step to refine the prediction by removing the parts which significantly overlap with others.

\vspace{2mm}
\noindent \textbf{Results and Analysis.} We show the results of our method for three object categories -- chairs, aeroplanes and animals in \figref{predFig}. We observe that the predictions successfully capture the coarse structure and are consistent across objects. The results indicate that the we can handle structural variations within a category \eg the objects in the right side of \figref{predFig} have a different structure than those on the left which occur more commonly in the dataset. 

We visualize in \figref{trainingErr} the training error across iterations. We observe that in the initial training stage (up to 20000 iterations), the loss rapidly decreases as the correct configuration is being learned. In the second stage of training, when we allow $p_m$ to be learned, the error initially increases -- this is because some primitives, encouraged by the parsimony reward, now start disappearing and the network eventually learns to use fewer primitives better.  Even though the reconstruction error in the initial stages is lower, the reconstructions using fewer primitives, are more parsimonious. This provides an insight regarding the tradeoff between representation parsimony  and reconstruction accuracy -- and that we should not judge the former by the latter.


\section{Applications}
\seclabel{applications}
\begin{figure*}[t!]
\centering
\includegraphics[width=1.0\textwidth]{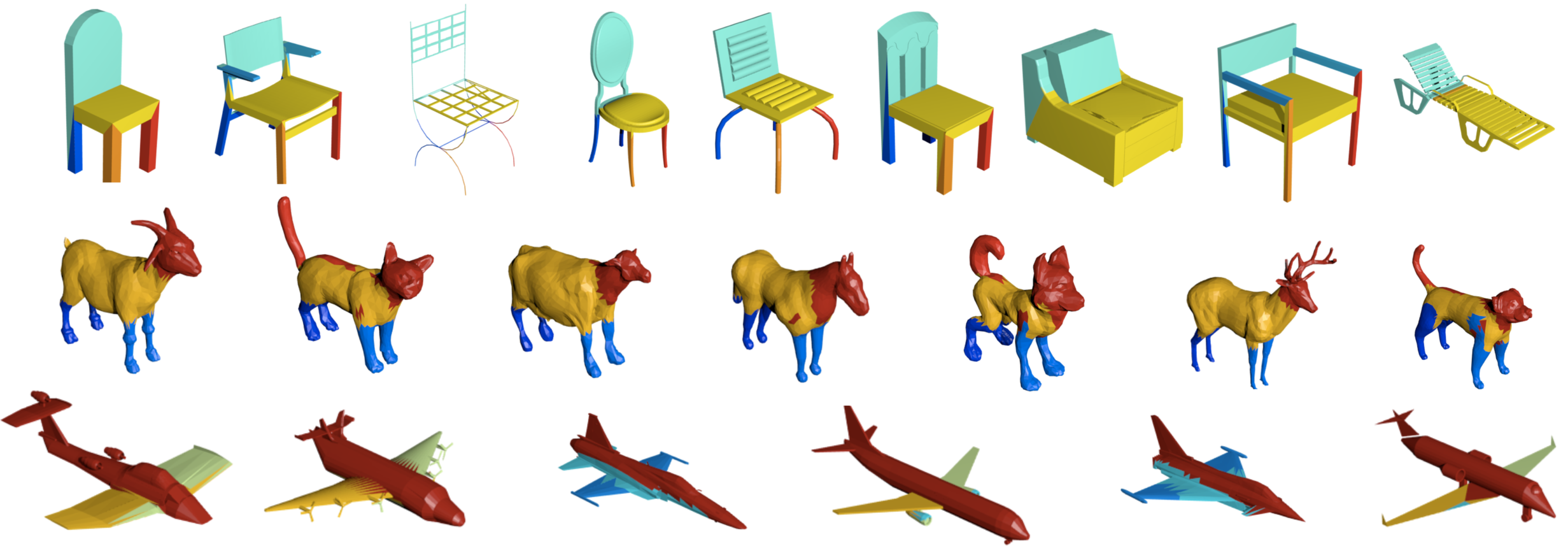}
\caption{Projection of the predicted primitives onto the original shape. We assign each point $p$ in the original shape to the corresponding primitive with lowest distance field $\mathcal{C}(p, \bar{P}_m)$. We visualize the parsing by coloring each point according to the assigned primitive. We see that similar parts \eg aeroplane wings, chair seat, \etc are consistently colored.}
\figlabel{parsing}
\end{figure*}

\begin{figure*}[t!]
\centering
\includegraphics[width=1.0\textwidth]{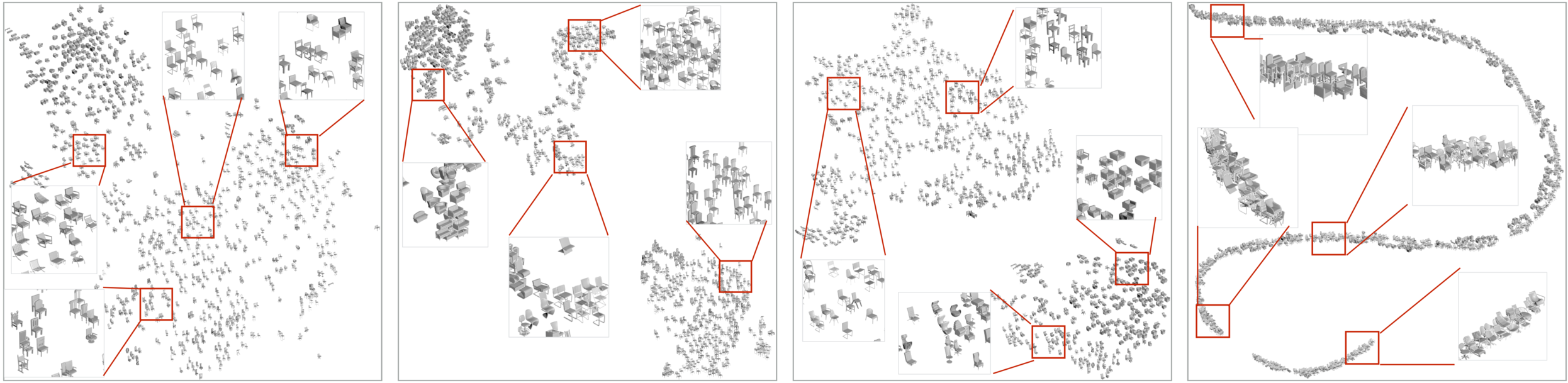}
\caption{Embeddings computed using various distance measures - a) Voxel IoU based distance b) Ours (all primitives) c) Ours (chair back, seat primitives) d) Ours (chair back orientation). While the IoU based embedding conflates chairs different fine level structure (\eg with/without handles), our embedding using all primitives encodes them separately. Additionally, unlike common shape representations, our inferred abstractions give us control over similarity measures -- we can choose to consider only specific primitives if required \eg chair back and seat which, as expected, results in ignoring existence of chair handles. We can also focus on specific properties \eg chair back orientation and observe a 1D manifold emerge in this scenario.  See appendix for high-resolution images.}
\figlabel{embedding}
\end{figure*}

We observe in \figref{fig1} and \figref{predFig} that the inferred representations are consistent across a category -- chair seat is explained consistently using the same primitive. They are also descriptive of the underlying shape and are, by construction, interpretable. Therefore, our framework allows us to automatically discover descriptive, consistent and interpretable shape abstractions using a collection of 3D models. By virtue of these properties, our representation can enable several applications related to shape similarity, part discovery, perception and shape manipulation.

\subsection{Unsupervised Parsing and Correspondence}
\seclabel{parsing}
The learned primitive decomposition is useful for obtaining part-level correspondences across instances. Since we use a common network across an object category, simple and consistent solutions are preferred to explain the data \ie the same primitive explains the chair back across the category. We can leverage this observation to extract correspondences across the category by assigning labels to points according to the primitive that explains them -- we assign each point to the primitive that has the lowest $\mathcal{C}(p, \bar{P}_m)$, giving preference to larger primitives to break ties. We therefore obtain a consistent labelling of all points across instances using the predicted primitive decomposition -- some examples are depicted in \figref{parsing}.

We also evaluate this parsing on the Shape COSEG \cite{wang2012active} dataset by measuring the accuracy using annotated ground-truth.
While the ground-truth only has 3 clusters (chair back, seat, legs), our method as well as previous unsupervised approaches~\cite{sidi2011unsupervised,wang2012active} cluster shapes into a larger number of partitions (number of primitives in our case) and assign each partition a ground-truth label to evaluate. We obtain a mean accuracy of 89.0$\%$ whereas~\cite{sidi2011unsupervised} reports 78.6$\%$ and 84.8$\%$ accuracy  with initial and refined parsings respectively\footnote{Unfortunately, we found that~\cite{sidi2011unsupervised} used a preliminary version of the Shape COSEG dataset~\cite{wang2012active}. We were unable to obtain this preliminary version, therefore the results are not exactly comparable. The algorithm in ~\cite{wang2012active} does use the current dataset but reports no quantitative results.}. See appendix for qualitative results. 

\subsection{Interpretable Shape Similarity}
\seclabel{similarity}
The trained CNN of our shape assembler maps every 3D shape to corresponding primitive parameters $\{(z_m,q_m,t_m)\}$. These parameters succinctly capture the geometry of the underlying object. We find that a simple euclidean distance in the embedding space is a reliable measure of shape similarity. We use this distance to compute a t-sne~\cite{maaten2008visualizing} embedding of shapes and visualize 1000 random instances in \figref{embedding} . We observe that the automatically discovered structure captures similarity better than a simple voxel IoU based metric and that clusters correspond to natural sub-categories \eg sofa \etc.

One aspect unique to our approach is that the shape embedding is interpretable and instead of using primitive parameters for all parts, we can modify the distance measure to focus on specifics of interest for the application. As an example, we show the resulting t-sne embedding if only 2 primitives, which correspond to back and seat, are used to compute the distance across shapes. We observe that the embedding reflects the desired similarity \eg unlike in the case of using all primitives to measure shape similarity, chairs with and without handles are now embedded together. We also compute the embedding for the distance measure which only measures the difference in the orientation $(q_m)$ for a specific part (chair back) and observe that this is a 1D manifold with the tilt increasing as we traverse it. Therefore, unlike common shape representations, our inferred abstractions give us control over similarity measures.


\begin{figure}[htb!]
\centering
\includegraphics[width=0.45\textwidth]{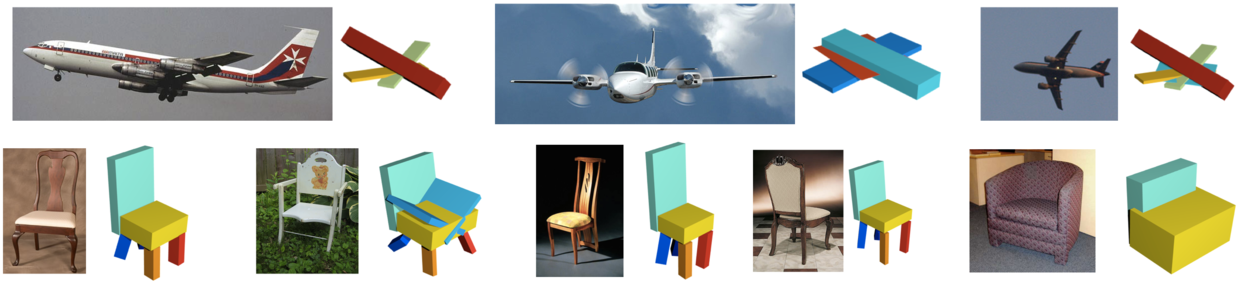}
\caption{Inferred abstractions using real image inputs.}
\figlabel{imgpred}
\end{figure}

\begin{figure}[htb!]
\centering
\includegraphics[width=0.45\textwidth]{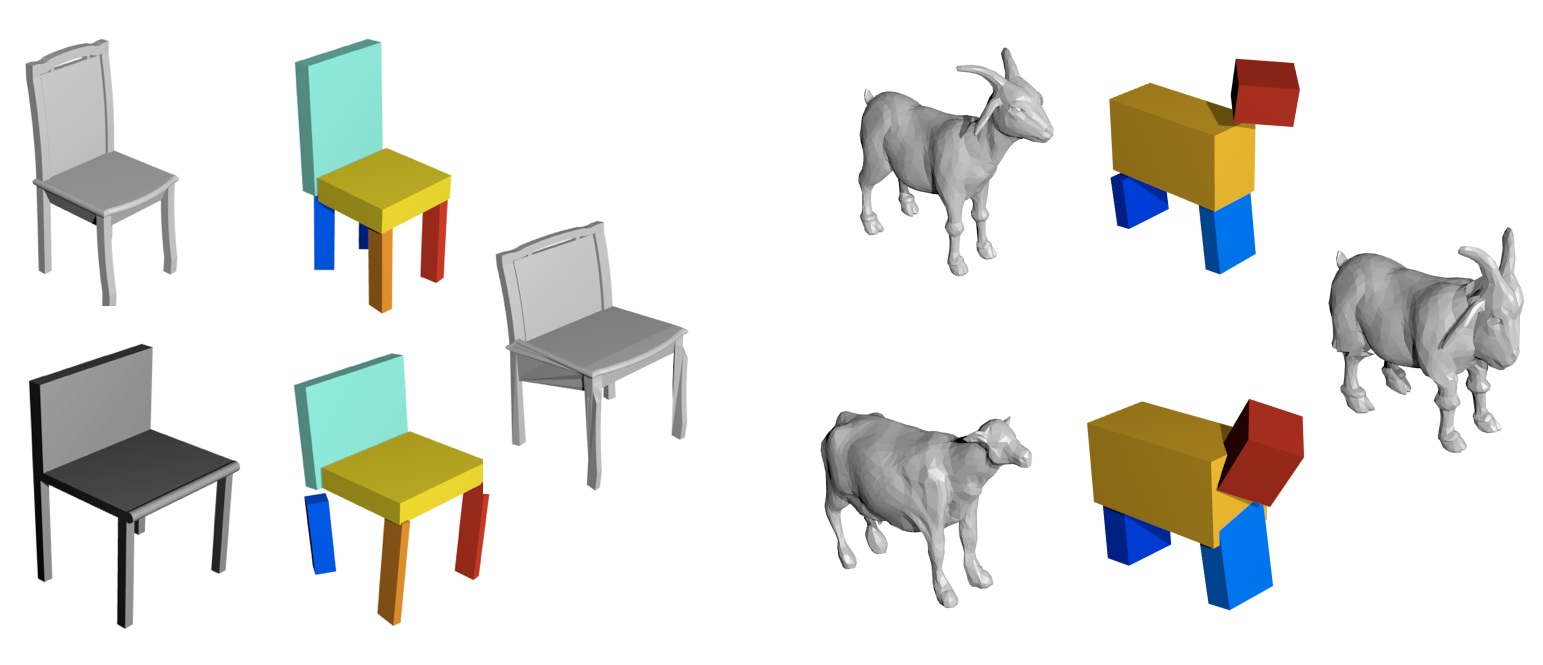}
\caption{We deform the source mesh (top) to have a shape similar to the target mesh (bottom) by using the inferred primitive representation. Each source mesh point is assigned a local coordinate in the closest primitive's frame. A deformation of the primitives from the source to target configuration induces a deformed mesh (shown on right).}
\figlabel{manipulation}
\end{figure}

\subsection{Image based Abstraction}
\seclabel{imgpred}
Given our trained model $h_\theta$ which infers primitive representation using volume inputs, we can train an image based prediction model $g_{\theta '}$. We obtain volume-image pairs $(V_i, I_i)$ by rendering ShapeNet models with random lighting and background (as suggested in \cite{su2015render}) and train the image based network to mimic the volume based network's predictions \ie we train $g_{\theta '}$ to minimize $\|h_\theta(V_i) -  g_{\theta '}(I_i) \|^2$. This distillation  technique~\cite{hinton2015distilling} for using paired data to train a model for predicting outputs similar to a pre-trained CNN is common~\cite{Gupta_2016_CVPR} and has previously also been used for learning shape embeddings~\cite {Girdhar16b}. We find that we can successfully apply this to our scenario and learn an image-based prediction model that outputs the abstraction of the underlying shape given a single image. We show some results in \figref{imgpred}. This demonstrates that one can learn to predict shape abstractions  using varying inputs and this might enable applications in robotics settings where such inference might help in grasping, planning \etc.

\subsection{Shape Manipulation}
\seclabel{manipulation}
The inferred primitive based shape abstractions can be used as a skeleton to guide manipulation of the underlying objects. We can assign each mesh point a local coordinate in the frame of its corresponding primitive (as computed in \secref{parsing}). A rotation, translation or scaling of the corresponding primitive can thereby induce a change in the global coordinates of the associated mesh points. We show some examples in \figref{manipulation} where we deform a source mesh to have a similar configuration as a target mesh. While the transformation used in this example is defined using a target mesh, one can also use our representation for other transformation \eg making the legs longer or tilting the back \etc.

\section{Conclusion}

In this work, we take an unsupervised, data-driven approach to explain visual information in terms of simpler primitives.  Taking inspiration from the classic work on generalized cylinders \cite{binford1971} and geons \cite{biederman1987recognition}, we too argue that any visual explanation must be in terms of 3D volumetric entities, not 2D pixel patches.  However, unlike the earlier work in this area we firmly believe in being data-driven and letting the data itself discover the best representation.

We demonstrated the applicability of data-driven 3D understanding of the visual world in a very simple setting -- that of explaining objects from cuboidal primitives. This merely represents the first steps towards the goal of generating parsimonious descriptions of the visual input and hope that this will motivate further efforts, including the use of a wider catalogue of basic parametrized primitives, to understand the underlying 3D structure of the world.

\section*{Acknowledgements}
We thank Saurabh Gupta and David Fouhey  for insightful discussions.
This work was supported in part by Intel/NSF Visual and Experiential Computing award IIS-1539099, NSF Award IIS-1212798, and the Berkeley Fellowship to ST. We gratefully acknowledge NVIDIA corporation for the donation of Tesla GPUs used for this research.

{\small
\bibliographystyle{ieee}
\bibliography{cvpr17abstraction}

\begin{thebibliography}{10}\itemsep=-1pt

\bibitem{biederman1987recognition}
I.~Biederman.
\newblock Recognition-by-components: a theory of human image understanding.
\newblock {\em Psychological review}, 1987.

\bibitem{binford1971}
T.~O. Binford.
\newblock Visual perception by computer.
\newblock In {\em IEEE Conference on Systems and Control}, 1971.

\bibitem{shapenet2015}
A.~X. Chang, T.~Funkhouser, L.~Guibas, P.~Hanrahan, Q.~Huang, Z.~Li,
  S.~Savarese, M.~Savva, S.~Song, H.~Su, J.~Xiao, L.~Yi, and F.~Yu.
\newblock {ShapeNet: An Information-Rich 3D Model Repository}.
\newblock Technical Report arXiv:1512.03012 [cs.GR], 2015.

\bibitem{cheung2014discovering}
B.~Cheung, J.~A. Livezey, A.~K. Bansal, and B.~A. Olshausen.
\newblock Discovering hidden factors of variation in deep networks.
\newblock {\em arXiv preprint arXiv:1412.6583}, 2014.

\bibitem{eslami2016attend}
S.~Eslami, N.~Heess, T.~Weber, Y.~Tassa, K.~Kavukcuoglu, and G.~E. Hinton.
\newblock Attend, infer, repeat: Fast scene understanding with generative
  models.
\newblock {\em arXiv preprint arXiv:1603.08575}.

\bibitem{faktor2013}
A.~Faktor and M.~Irani.
\newblock Co-segmentation by composition.
\newblock In {\em ICCV 2013}.

\bibitem{fakscm_metarep_sig14}
N.~Fish, M.~Averkiou, O.~van Kaick, O.~Sorkine-Hornung, D.~Cohen-Or, and N.~J.
  Mitra.
\newblock Meta-representation of shape families.
\newblock {\em Transactions on Graphics (SIGGRAPH)}, 2014.

\bibitem{Girdhar16b}
R.~Girdhar, D.~Fouhey, M.~Rodriguez, and A.~Gupta.
\newblock Learning a predictable and generative vector representation for
  objects.
\newblock In {\em ECCV}, 2016.

\bibitem{goodfellow2014generative}
I.~Goodfellow, J.~Pouget-Abadie, M.~Mirza, B.~Xu, D.~Warde-Farley, S.~Ozair,
  A.~Courville, and Y.~Bengio.
\newblock Generative adversarial nets.
\newblock In {\em NIPS}, 2014.

\bibitem{gregor2015draw}
K.~Gregor, I.~Danihelka, A.~Graves, and D.~Wierstra.
\newblock Draw: A recurrent neural network for image generation.
\newblock {\em arXiv preprint arXiv:1502.04623}, 2015.

\bibitem{gupta2010blocks}
A.~Gupta, A.~A. Efros, and M.~Hebert.
\newblock Blocks world revisited: Image understanding using qualitative
  geometry and mechanics.
\newblock In {\em ECCV}. 2010.

\bibitem{Gupta_2016_CVPR}
S.~Gupta, J.~Hoffman, and J.~Malik.
\newblock Cross modal distillation for supervision transfer.
\newblock In {\em CVPR}, June 2016.

\bibitem{hinton2015distilling}
G.~Hinton, O.~Vinyals, and J.~Dean.
\newblock Distilling the knowledge in a neural network.
\newblock {\em arXiv preprint arXiv:1503.02531}, 2015.

\bibitem{huang2015efficient}
J.~Huang and K.~Murphy.
\newblock Efficient inference in occlusion-aware generative models of images.
\newblock {\em arXiv preprint arXiv:1511.06362}, 2015.

\bibitem{huang2011joint}
Q.~Huang, V.~Koltun, and L.~Guibas.
\newblock Joint shape segmentation with linear programming.
\newblock In {\em ACM Transactions on Graphics (TOG)}. ACM, 2011.

\bibitem{huang2015single}
Q.~Huang, H.~Wang, and V.~Koltun.
\newblock Single-view reconstruction via joint analysis of image and shape
  collections.
\newblock {\em ACM Transactions on Graphics (TOG)}, 2015.

\bibitem{SceneCollaging}
P.~Isola and C.~Liu.
\newblock Scene collaging: Analysis and synthesis of natural images with
  semantic layers.
\newblock In {\em ICCV}, 2013.

\bibitem{kalogerakis2012}
E.~Kalogerakis, S.~Chaudhuri, D.~Koller, and V.~Koltun.
\newblock A probabilistic model for component-based shape synthesis.
\newblock {\em ACM Transactions on Graphics (TOG)}, 31(4):55, 2012.

\bibitem{shapesKarTCM15}
A.~Kar, S.~Tulsiani, J.~Carreira, and J.~Malik.
\newblock Category-specific object reconstruction from a single image.
\newblock In {\em CVPR}, 2015.

\bibitem{adam}
D.~Kingma and J.~Ba.
\newblock Adam: A method for stochastic optimization.
\newblock {\em arXiv preprint arXiv:1412.6980}.

\bibitem{kingma2013auto}
D.~P. Kingma and M.~Welling.
\newblock Auto-encoding variational bayes.
\newblock {\em arXiv preprint arXiv:1312.6114}, 2013.

\bibitem{kulkarni2015deep}
T.~D. Kulkarni, W.~Whitney, P.~Kohli, and J.~B. Tenenbaum.
\newblock Deep convolutional inverse graphics network.
\newblock {\em arXiv preprint arXiv:1503.03167}, 2015.

\bibitem{li2015joint}
Y.~Li, H.~Su, C.~R. Qi, N.~Fish, D.~Cohen-Or, and L.~J. Guibas.
\newblock Joint embeddings of shapes and images via cnn image purification.
\newblock {\em TOG 2015}.

\bibitem{limparsing}
J.~J. Lim, H.~Pirsiavash, and A.~Torralba.
\newblock Parsing ikea objects: Fine pose estimation.
\newblock In {\em ICCV 2013}.

\bibitem{maaten2008visualizing}
L.~v.~d. Maaten and G.~Hinton.
\newblock Visualizing data using t-sne.
\newblock {\em JMLR}, 2008.

\bibitem{radford2015unsupervised}
A.~Radford, L.~Metz, and S.~Chintala.
\newblock Unsupervised representation learning with deep convolutional
  generative adversarial networks.
\newblock {\em arXiv preprint arXiv:1511.06434}, 2015.

\bibitem{roberts1963machine}
L.~G. Roberts.
\newblock {\em Machine Perception of Three-Dimensional Solids}.
\newblock PhD thesis, MIT, 1963.

\bibitem{rubinstein2013}
M.~Rubinstein, A.~Joulin, J.~Kopf, and C.~Liu.
\newblock Unsupervised joint object discovery and segmentation in internet
  images.
\newblock In {\em CVPR 2013}.

\bibitem{russell2009segmenting}
B.~Russell, A.~Efros, J.~Sivic, B.~Freeman, and A.~Zisserman.
\newblock Segmenting scenes by matching image composites.
\newblock In {\em NIPS}, 2009.

\bibitem{russell2006using}
B.~C. Russell, A.~A. Efros, J.~Sivic, W.~T. Freeman, and A.~Zisserman.
\newblock Using multiple segmentations to discover objects and their extent in
  image collections.
\newblock In {\em CVPR 2006}.

\bibitem{sidi2011unsupervised}
O.~Sidi, O.~van Kaick, Y.~Kleiman, H.~Zhang, and D.~Cohen-Or.
\newblock Unsupervised co-segmentation of a set of shapes via descriptor-space
  spectral clustering.
\newblock {\em ACM Trans. on Graphics (Proc. SIGGRAPH Asia)}, 2011.

\bibitem{su2015render}
H.~Su, C.~R. Qi, Y.~Li, and L.~J. Guibas.
\newblock Render for cnn: Viewpoint estimation in images using cnns trained
  with rendered 3d model views.
\newblock In {\em ICCV}, 2015.

\bibitem{sung2015data}
M.~Sung, V.~G. Kim, R.~Angst, and L.~Guibas.
\newblock Data-driven structural priors for shape completion.
\newblock {\em ACM Transactions on Graphics (TOG)}, 2015.

\bibitem{van2015part}
A.~van~den Hengel, C.~Russell, A.~Dick, J.~Bastian, D.~Pooley, L.~Fleming, and
  L.~Agapito.
\newblock Part-based modelling of compound scenes from images.
\newblock In {\em CVPR 2015}.

\bibitem{vicente2014reconstructing}
S.~Vicente, J.~Carreira, L.~Agapito, and J.~Batista.
\newblock Reconstructing pascal voc.
\newblock In {\em CVPR}, 2014.

\bibitem{wang2012active}
Y.~Wang, S.~Asafi, O.~van Kaick, H.~Zhang, D.~Cohen-Or, and B.~Chen.
\newblock Active co-analysis of a set of shapes.
\newblock {\em ACM Transactions on Graphics (TOG)}, 2012.

\bibitem{williams1992simple}
R.~J. Williams.
\newblock Simple statistical gradient-following algorithms for connectionist
  reinforcement learning.
\newblock {\em Machine learning}, 1992.

\bibitem{pascal3d}
Y.~Xiang, R.~Mottaghi, and S.~Savarese.
\newblock Beyond pascal: A benchmark for 3d object detection in the wild.
\newblock In {\em WACV}, 2014.

\bibitem{yumer2012co}
M.~E. Yumer and L.~B. Kara.
\newblock Co-abstraction of shape collections.
\newblock {\em ACM Transactions on Graphics (TOG)}, 2012.

\bibitem{yumer2014co}
M.~E. Yumer and L.~B. Kara.
\newblock Co-constrained handles for deformation in shape collections.
\newblock {\em ACM Transactions on Graphics (TOG)}, 2014.

\bibitem{zcam_partArrangement_eg14}
Y.~Zheng, D.~Cohen-Or, M.~Averkiou, and N.~J. Mitra.
\newblock Recurring part arrangements in shape collections.
\newblock {\em Computer Graphics Forum (Eurographics 2014)}, 2014.

\end{thebibliography}
}
\clearpage
\section*{Appendix}

\subsection*{A1. Consistency Loss via Differentiable Surface Sampling}
We first re-iterate the consistency loss which aims for the assembled shape to be subsumed by ground-truth object and does so by sampling points on the surface of each primitive and penalizing the squared distance field w.r.t the ground-truth object at the sampled points.

\begin{gather*}
L_2(\{(z_m, q_m, t_m)\},O) = \underset{m}{\sum} ~ \mathbb{E}_{p \sim S(\bar{P}_m)} \|  \mathcal{C}(p; O) \|^2
\eqlabel{consistencyLoss}
\end{gather*}

Recall that to sample a point $p$ on the surface of $\bar{P}_m$, one can equivalently sample $p'$ on the surface of the untransformed primitive $P_m$ and then rotate, translate $p'$ according to $(q_m, z_m)$.
\begin{gather*}
p \sim S(\bar{P}_m) \equiv \mathcal{T}(\mathcal{R}(p', q_m), t_m); p' \sim S(P_m)
\end{gather*}

Note that the untransformed primitive $P_m$ is an origin-centered cuboid parametrized by $z_m \equiv (w_m, h_m, d_m)$ -- the dimensions along the three canonical axes. We now aim to show that we can derive gradients for $z_m$ given gradients for a $p' \sim S(P_m)$. We do so using a re-parametrization trick where we aim to decouple the random sampling process from the parameters. The process of sampling from a cuboid's surface requires developing tow aspects. The first concerns how to sample from a particular face. The second required component is to decide which face to sample from -- we show that we can sample equally from all faces if we assign sample importance weights.

\vspace{2mm}
\noindent \textbf{Sampling Along a Face.}
Let us assume we want to to sample a point from a given cuboid face -- say the one on the plane $x=w$. The process of sampling a point on this face can be written as -
 \begin{gather*}
u_x = 1; u_y \sim [-1,1]; u_z \sim [-1,1]; \\
p' = (u_xw_m,u_yh_m,u_zd_m)
\end{gather*}
We can similarly define a sampling process for the other cuboid faces. Note that the coordinates of the sampled point $p'$ are linear the primitive parameters and that given the random coefficients $(u_x,u_y,u_z)$, it is straightforward to compute $\frac{\partial p'}{\partial z_m}$.

\vspace{2mm}
\noindent \textbf{Importance Weights per Sample.} To sample uniformly on the cuboid surface, we need to sample on each face with a probability proportional to its area. Unfortunately, it is unclear how this sampling process can be decoupled from the parameters $p_m$. Instead, we can simply sample equally from all faces and assign each sample an importance weight proportional to the area of the face it was sampled from. Under this implementation, the consistency loss as previously defined can be implemented as:
\begin{gather*}
L_2(\{(z_m, q_m, t_m)\},O) = \underset{m}{\sum} ~  \underset{p,w_p \sim S(\bar{P}_m)}{\sum} w_p \|  \mathcal{C}(p; O) \|^2
\eqlabel{consistencyLossImp}
\end{gather*}
Here $w_p$ is the importance weight of the corresponding sample and is proportional to the area of the face where it was drawn from. Note that $w_p$ is also differentiable w.r.t $z_m$, therefore making consistency loss described differentiable w.r.t the predicted primitive parameters.

\subsection*{A2. Gradient computation for primitive existence probabilities  $p_m$}
We provide more details on how we use the REINFORCE algorithm to compute gradients for the primitive existence probabilities $p_m$. We discuss a more general  implementation compared to the text -- we assume $p_m \equiv (p_m^0, p_m^1)$ represents the probability of the available choices for the primitive -- $p_m^0$ is the probability that the primitive does not exist and $p_m^1$ is the probability that it is a cuboid. Note that this can be modified to add other choices \eg a third choice that the primitive is a cylinder.

Let $l$ denote the loss $L(\{(z_m, q_m, t_m)\}, O)$ incurred for the particular sampling of $z_m^e$. Let $r$ denote the external parsimony reward obtained if a primitive is sampled as not existing. Using these, the gradient for the predicted probability $p_m$ is computed as -
\begin{gather*}
\frac{\partial L}{\partial p_m^i} =
\begin{dcases}
   \frac{l - \mathbbm{1}(z_m^e=0)r}{p_m^i} , & \text{if } z_m^e = i\\
   0,   & \text{otherwise}\\
\end{dcases}
\end{gather*}
It is typically also advised to subtract a `baseline' (mean reward in the ideal scenario) to reduce the variance of the estimated gradients so we incorporate it using $l-b$ instead of $l$ in the equation above where $b$ is the running average of the losses incurred.

\subsection*{A3. Architecture and Initialization}
\vspace{2mm}
\noindent \textbf{Architecture and Hyper-parameters.}
Our network has five 3D convolution layers (with ReLU) of kernel 3, padding 1, stride 2. The number of channels (initially 4) are doubled after every layer. These are followed by 2 fc layers (with ReLU) with 100 units each, followed by a final layer to predict primitive parameters. We use ADAM for optimization and choose M to be more than the double the number of expected parts. M=20 for chairs, but was reduced to 15, 12 for planes, animals (to reduce iteration time).

\vspace{2mm}
\noindent \textbf{Initialization.}
We initially bias $z_m$ s.t. primitives are small cubes  and  $p_m=0.9$~so every primitive is initially used but all other layer parameters/weights (and primitive positions, rotations \etc.) are initialized randomly (our training  discovers consistent solutions despite a random initialization because of commonalities in the data). 

\subsection*{A4. Visualization}

\vspace{2mm}
\noindent \textbf{Shape COSEG Results.} We visualize our parsings obtained on the `chairs' subset of the Shape COSEG data. \figref{scmesh} shows the meshes in the dataset and \figref{scabs} shows our inferred representations for these. \figref{scparse} visualizes the obtained unsupervised parsing of the original mesh. \figref{scgt} shows the annotated ground-truth labels and \figref{scpred} shows our predictions obtained by assigning each primitive a label from the available ground-truth labels.

\begin{figure*}[t!]
\centering
\includegraphics[width=0.95\textwidth]{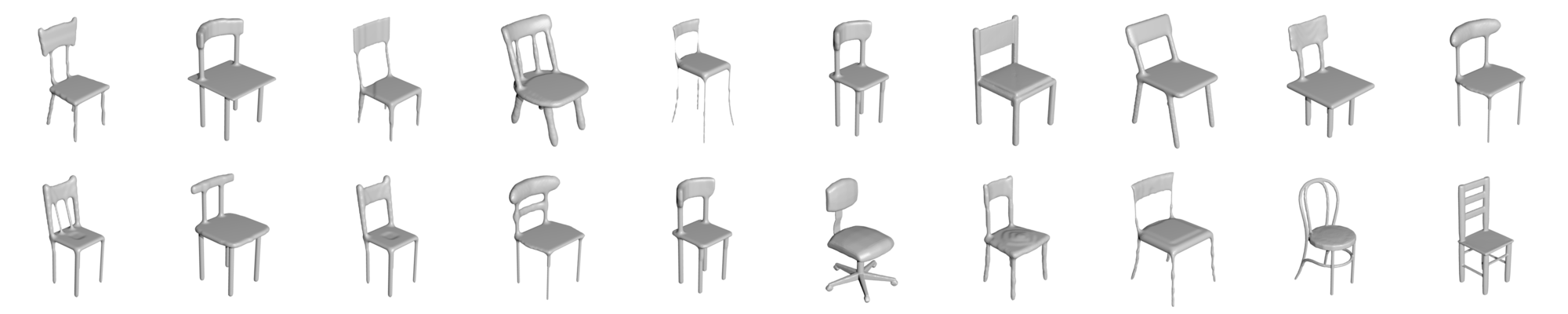}
\caption{Meshes in the Shape COSEG dataset.}
\figlabel{scmesh}
\end{figure*}

\begin{figure*}[t!]
\centering
\includegraphics[width=0.95\textwidth]{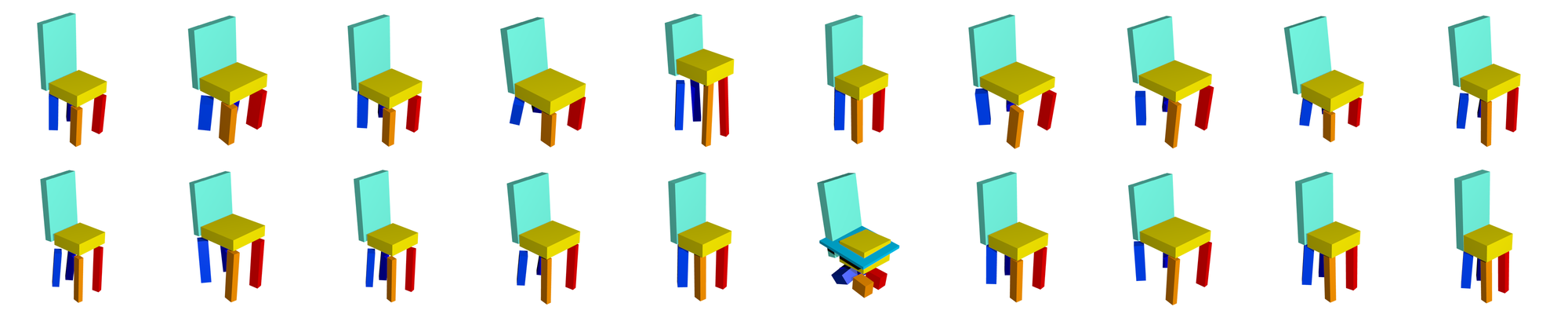}
\caption{Inferred primitive based representations using our learned CNN on ShapeNet.}
\figlabel{scabs}
\end{figure*}

\begin{figure*}[t!]
\centering
\includegraphics[width=0.95\textwidth]{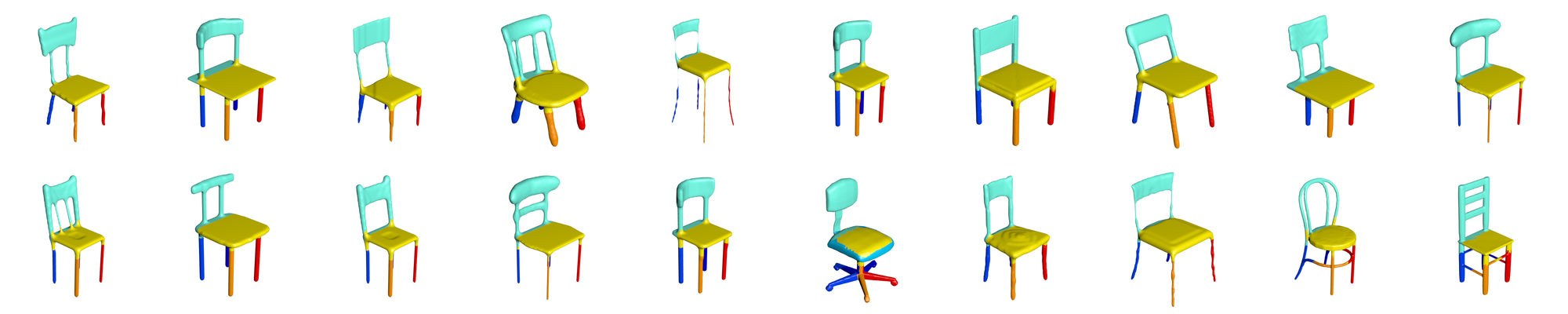}
\caption{Primitive labels projected onto the original mesh.}
\figlabel{scparse}
\end{figure*}

\begin{figure*}[t!]
\centering
\includegraphics[width=0.95\textwidth]{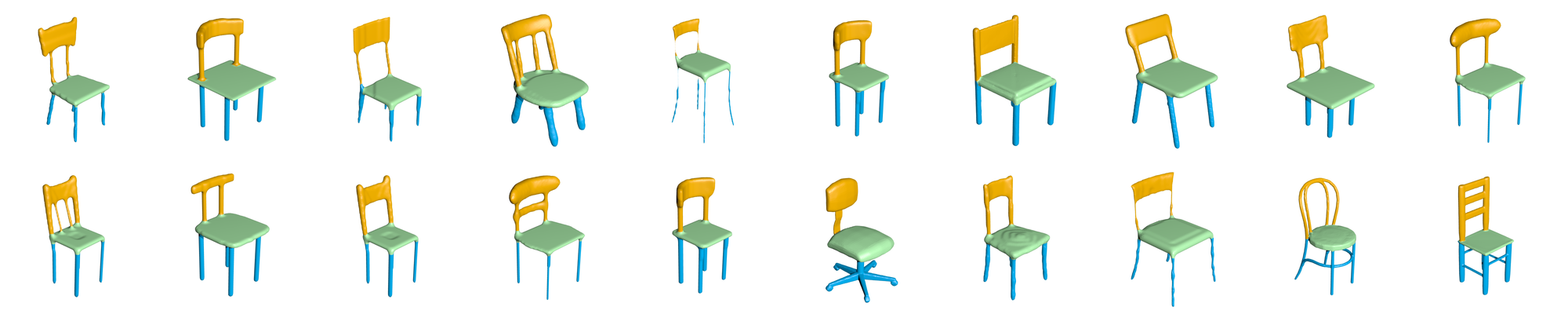}
\caption{Annotated Ground-truth part labels.}
\figlabel{scgt}
\end{figure*}

\begin{figure*}[t!]
\centering
\includegraphics[width=0.95\textwidth]{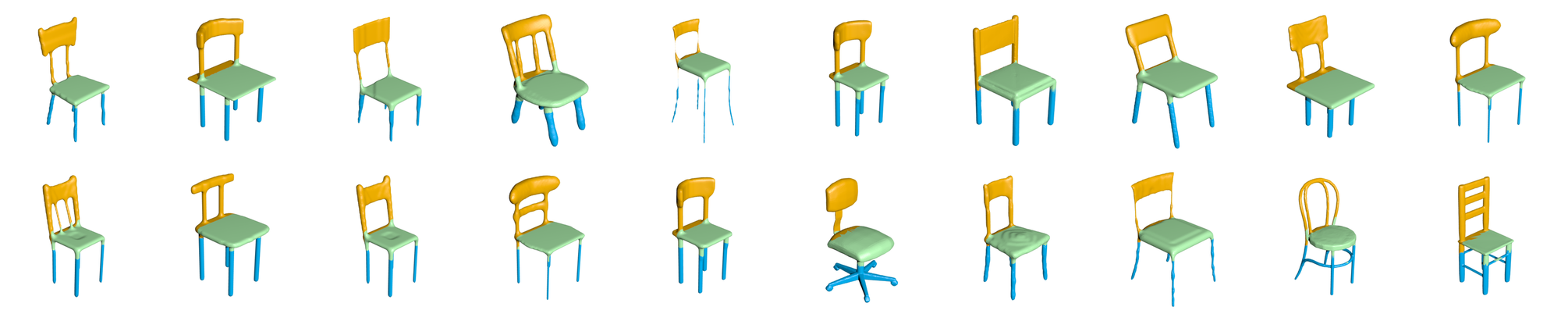}
\caption{Our inferred part labels on the Shape COSEG dataset, obtained by assigning each primitive to a ground-truth label.}
\figlabel{scpred}
\end{figure*}

\paragraph{Shape Embedding.}
We show in \figref{embedVox},  \figref{embedOurs1}, \figref{embedOurs2} and \figref{embedOurs3} the high-resolution images corresponding to the embeddings shown in the original text.
\begin{figure*}[t!]
\centering
\includegraphics[width=1.00\textwidth]{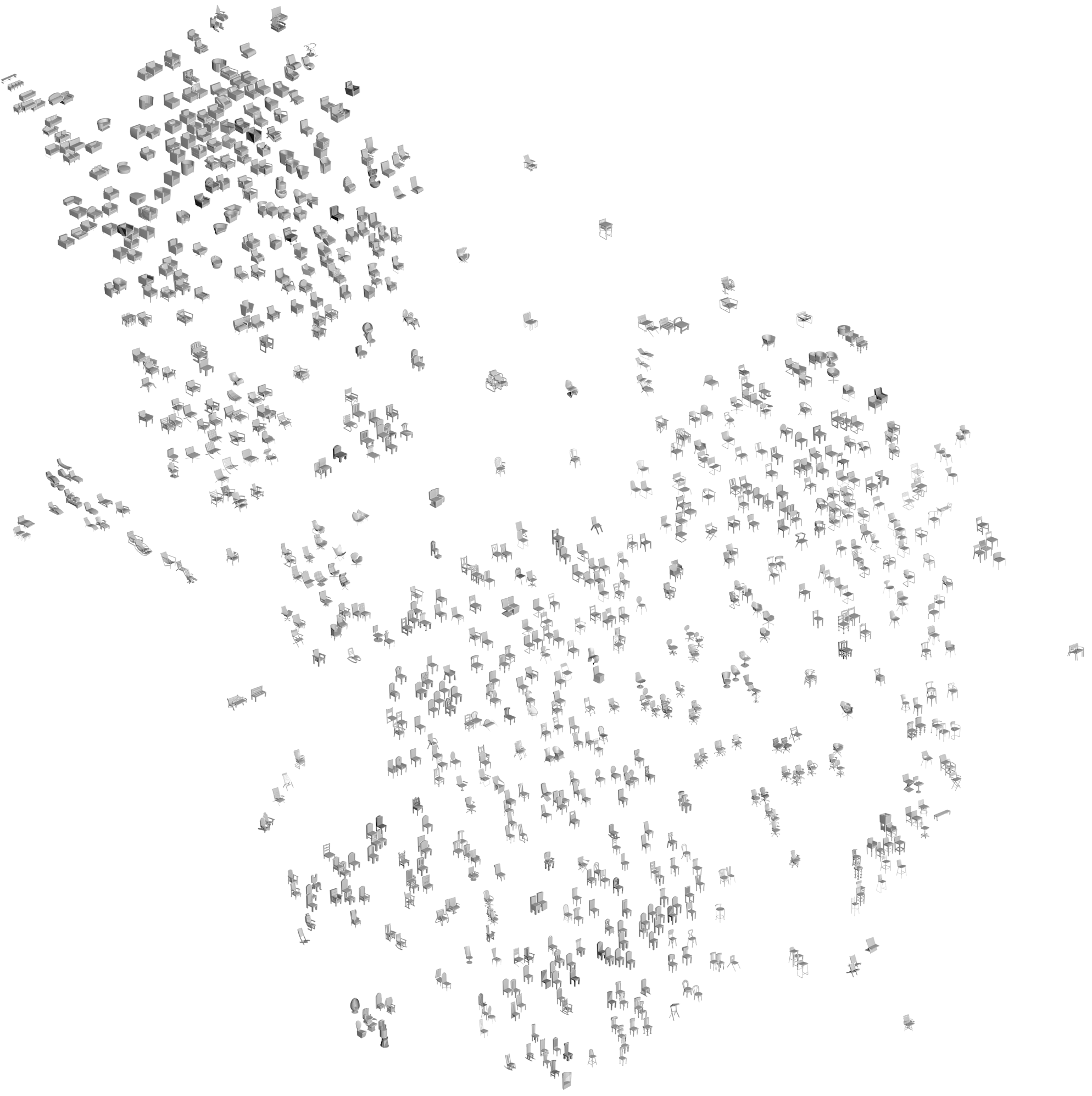}
\caption{Embedding using voxel IoU as distance metric.}
\figlabel{embedVox}
\end{figure*}

\begin{figure*}[t!]
\centering
\includegraphics[width=1.00\textwidth]{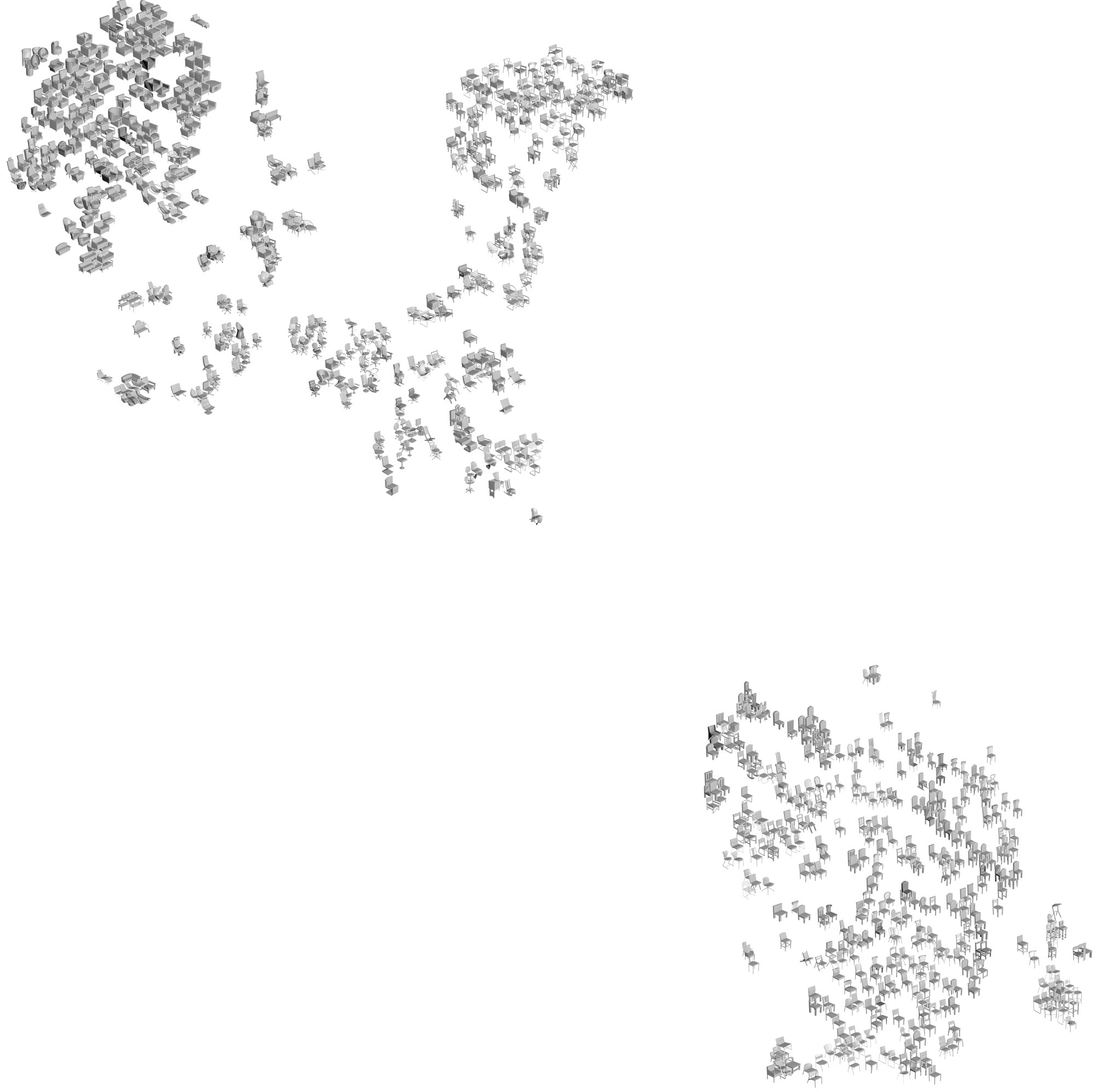}
\caption{Embedding using squared euclidean distance in primitive based representation space.}
\figlabel{embedOurs1}
\end{figure*}

\begin{figure*}[t!]
\centering
\includegraphics[width=1.00\textwidth]{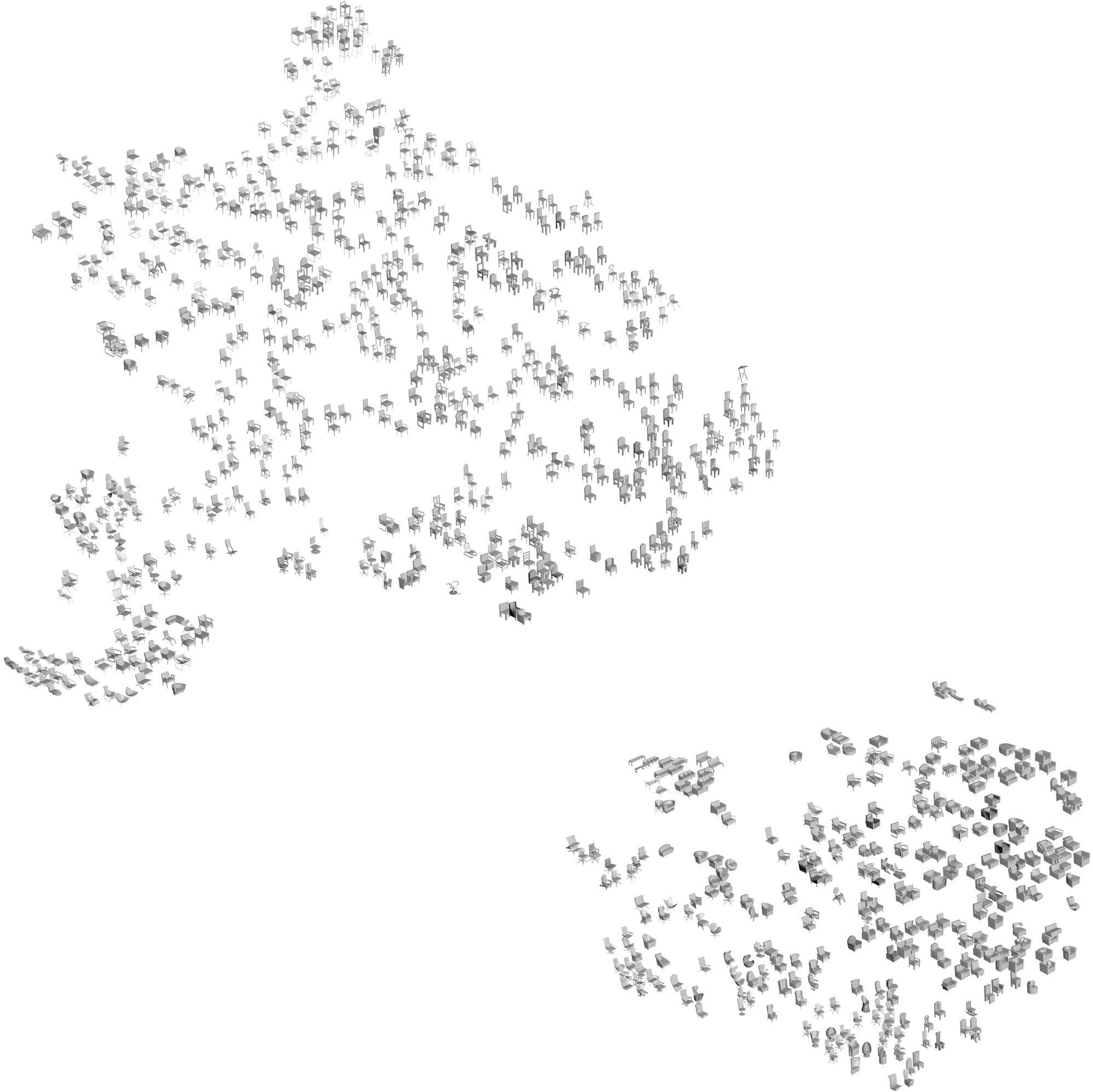}
\caption{Embedding using configurations of two specific primitives (chair back, seat) to compute the distances.}
\figlabel{embedOurs2}
\end{figure*}

\begin{figure*}[t!]
\centering
\includegraphics[width=1.00\textwidth]{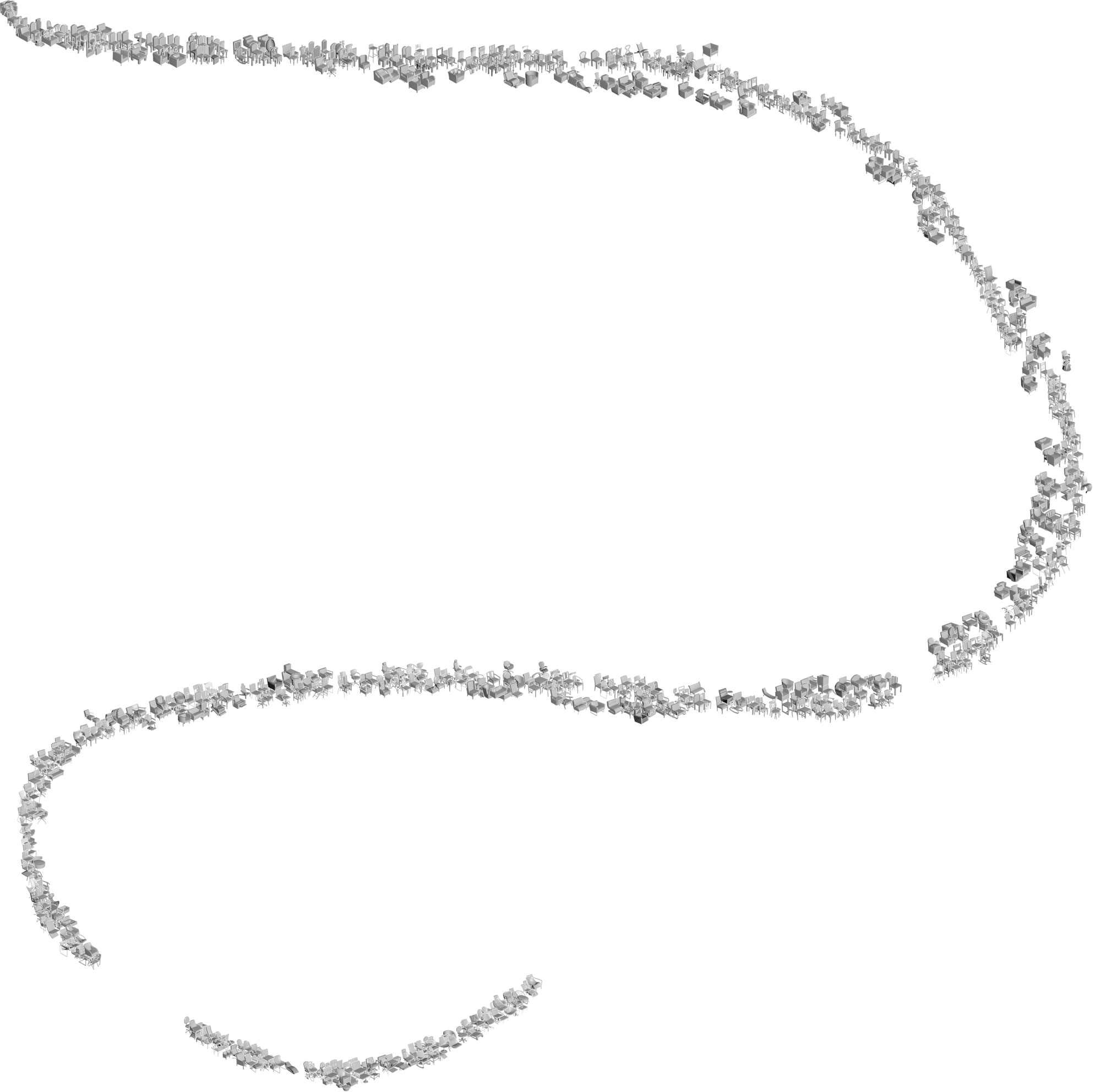}
\caption{Embedding using the orientation of a specific primitive (chair back) to compute distances.}
\figlabel{embedOurs3}
\end{figure*}

\end{document}